\documentclass{article} % For LaTeX2e
\usepackage{iclr2024_conference,times}

\usepackage{amsmath,amsfonts,bm}

\def\eqref#1{equation~\ref{#1}}
\def\1{\bm{1}}

\DeclareMathAlphabet{\mathsfit}{\encodingdefault}{\sfdefault}{m}{sl}
\SetMathAlphabet{\mathsfit}{bold}{\encodingdefault}{\sfdefault}{bx}{n}

\usepackage{hyperref}
\usepackage{url}
\usepackage{hyperref}
\usepackage{caption}
\usepackage{subcaption}
\usepackage{cleveref}
\usepackage{amsthm, thmtools}
\usepackage{thm-restate}
\usepackage{bbm}
\usepackage{amssymb}

\usepackage{graphicx}
\usepackage{booktabs}
\usepackage{multirow}
\usepackage{wrapfig}
\usepackage{algorithm}
\usepackage{algpseudocode}
\algrenewcommand\algorithmicrequire{\textbf{Input:}}
\algrenewcommand\algorithmicensure{\textbf{Output:}}

\newcommand{\Ebb}{\mathop{\mathbb{E}}}
\newcommand{\Rbb}{\mathbb{R}}
\newcommand{\Lcal}{\mathcal{L}}
\newcommand{\Ncal}{\mathcal{N}}
\newcommand{\ind}{\mathop{\mathbbm{1}}}
\newcommand{\ropt}{r_{\operatorname{opt}}}
\newcommand{\randf}{f_{\operatorname{rand}}}
\newcommand{\acc}{\operatorname{Acc}}

\newtheorem{assumption}{Assumption}
\newtheorem{definition}{Definition}

\title{Understanding the Robustness of Randomized Feature Defense Against Query-Based Adversarial Attacks}

\author{Quang H. Nguyen$^1$, Yingjie Lao$^2$, Tung Pham$^3$, Kok-Seng Wong$^1$, Khoa D. Doan$^1$ \\
$^1$College of Engineering and Computer Science, VinUniversity, Vietnam\\
$^2$Electrical and Computer Engineering,  Clemson University, USA\\
$^3$VinAI Research\\
\texttt{quang.nh@vinuni.edu.vn, ylao@clemson.edu, v.tungph4@vinai.io}\\ \texttt{wong.ks@vinuni.edu.vn, khoa.dd@vinuni.edu.vn} \\
}

\arxivtrue
\iclrfinalcopy % Uncomment for camera-ready version, but NOT for submission.
\begin{document}

\maketitle

\begin{abstract}
Recent works have shown that deep neural networks are vulnerable to adversarial examples that find samples close to the original image but can make the model misclassify. Even with access only to the model's output, an attacker can employ black-box attacks to generate such adversarial examples. In this work, we propose a simple and lightweight defense against black-box attacks by adding random noise to hidden features at intermediate layers of the model at inference time. Our theoretical analysis confirms that this method effectively enhances the model's resilience against both score-based and decision-based black-box attacks. Importantly, our defense does not necessitate adversarial training and has minimal impact on accuracy, rendering it applicable to any pre-trained model. Our analysis also reveals the significance of selectively adding noise to different parts of the model based on the gradient of the adversarial objective function, which can be varied during the attack. We demonstrate the robustness of our defense against multiple black-box attacks through extensive empirical experiments involving diverse models with various architectures.
\end{abstract}

\section{Introduction}
Modern deep neural networks have demonstrated remarkable performance in various complex tasks, including image classification and face recognition, among others. However, prior works have pointed out that deep learning models are sensitive to small changes in the input and can be fooled by carefully chosen and imperceptible perturbations~\cite{szegedy2013intriguing, goodfellow2014explaining, papernot2016limitations, madry2017towards}. These adversarial attacks can be generally classified into white-box and black-box attacks. In a white-box setting, strong attacks such as Projected Gradient Descent (PGD)~\cite{madry2017towards} can generate effective adversarial examples by levering the information inside the model. However, in practical scenarios such as machine learning as a service (MLaas), the well-trained models and the training datasets are often inaccessible to the users, especially in the era of large models. Hence, query-based black-box attacks become the primary threats in most real-world applications, where the adversary is assumed to have no knowledge of the model architecture and parameters.  

This paper proposes a lightweight, plug-and-play defensive method that can significantly decrease the success rate of query-based black-box attacks, including both score-based and decision-based attacks~\cite{ilyas2018black, ilyas2019prior, andriushchenko2020square, guo2019simple,al2020sign, liu2019signsgd, chen2020rays, Chen2020boosting, rahmati2020geoda}. Adversarial examples generated through query-based attacks involve iterative procedures that rely on either local search techniques involving small incremental modifications to the input or optimization methods leveraging estimated gradients of the adversary's loss concerning the input. However, the process of requesting numerous queries is time-consuming and sometimes may raise suspicions with the presence of multiple similar queries. Hence, the objective of defense is to perplex the adversary when attempting to generate adversarial examples. Our proposed method accomplishes this by introducing noise into the feature space. Unlike previous randomized defense approaches that solely rely on empirical evaluations to showcase effectiveness, this paper provides both theoretical analysis and empirical evidence to demonstrate improved robustness. Our analysis also highlights the importance of strategically introducing noise to specific components of the model based on the gradient of the adversarial objective function, which can be dynamically adjusted throughout the attack process.

Our contributions can be summarized as follows:
\begin{itemize}
    \item We investigate the impact of randomized perturbations in the feature space and its connection to the robustness of the model to black-box attacks. 
    \item We design a simple yet effective and lightweight defense strategy that hampers the attacker's ability to approximate the direction toward adversarial samples. As a result, the success rate of the attacks is significantly reduced.
    \item We extensively evaluate our approach through experiments on both score-based and decision-based attacks. The results validate our analysis and demonstrate that our method enhances the robustness of the randomized model against query-based attacks.
\end{itemize}

\section{Related Works}

\subsection{Adversarial Attacks}
Extensive research has been conducted on white-box attacks, focusing on the generation of adversarial examples when the attacker possesses complete access to the target model. Over the years, various notable methods have emerged as representative approaches in this field, including fast gradient sign method (FGSM)~\cite{goodfellow2014explaining}, Jacobian-based saliency Map Attack (JSMA)~\cite{papernot2016crafting}, and PGD~\cite{madry2017towards}.

In contrast to white-box attacks, the black-box scenario assumes that the attacker lacks access to the target model, making it a more challenging situation. However, this is also a more realistic setting in real-world applications where the adversary would not have access to the model parameters. One approach in black-box attacks involves utilizing white-box techniques on substitute models to create adversarial examples, which can subsequently be applied to black-box target models~\cite{papernot2017practical}.  However, the effectiveness of transfer-based attacks can vary significantly due to several practical factors, such as the initial training conditions, model hyperparameters, and constraints involved in generating adversarial samples~\cite{chen2017zoo}. This paper focuses on the defense against query-based attacks instead.

% intuition behind query based
% Some typical attacks
% Some typical defenses

\subsection{Query-based Black-box Attacks} Query-based attacks can be largely divided into score-based attacks and decision-based attacks, based on the accessible model output information. Score-based attacks leverage the output probability or logit of the targeted model, allowing the attacker to manipulate the scores associated with different classes. On the other hand, decision-based queries provide the attacker with hard labels, restricting the access to only the final predictions without any probability or confidence values. 

We list the query-based attacks used in this paper below:

\textbf{Natural Evolutionary Strategies (NES)~\cite{ilyas2018black}} is one of the first query-based attacks that use natural evolutional strategies to estimate the gradient of the model with respect to an image $x$. By exploring the queries surrounding $x$, NES effectively gauges the model's gradient, enabling it to probe and gain insights into the model's behavior.

\textbf{SignHunt~\cite{al2020sign}} is another score-based attack, which flips the sign of the perturbation based on the sign of the estimated gradient to improve the query efficiency.

\textbf{Square attack~\cite{andriushchenko2020square}} is a type of score-based attack that differs from gradient approximation techniques. Instead, it employs random search to update square-shaped regions located at random positions within the images. This approach avoids relying on gradient information and introduces a localized square modification to the image.

\textbf{RayS~\cite{chen2020rays}} is a decision-based attack that solves a discrete problem to find the direction with the smallest distance to the decision boundary while using a fast check step to avoid unnecessary searches.

\textbf{SignFlip~\cite{Chen2020boosting}} is an $\ell^{\infty}$ decision based attack that alternately projects the perturbation to a smaller $\ell^{\infty}$ ball and flips the sign of some randomly selected entries in the perturbation.

%list
\subsection{Defensive Methods against Query-based Attacks}
In the recent literature, several defensive solutions have been proposed to counter adversarial examples. One such solution involves the detection of malicious queries by comparing them with previously observed normal queries~\cite{chen2020stateful,li2022blacklight,pang2020advmind}. This approach aims to identify anomalous patterns in queries and flag them as potential adversarial examples. Additionally, adversarial training has also been utilized to enhance the model's robustness~\cite{cohen2019certified, wang2020certifying, sinha2017certifiable, zhang2020black}. Adversarial training involves training the model on both regular and adversarial examples to improve its ability to withstand adversarial attacks. However, it is computationally expensive, especially when dealing with large and complex datasets. In some cases, adversarial training may also inadvertently harm the model's overall performance. 

In contrast, this paper focuses on approaches that involve incorporating noise or randomness into the model, thereby providing the adversary with distorted information. The underlying intuition behind these defense mechanisms is to deceive the attacker by introducing perturbations in the model's prediction process. By altering certain signals, the defenses aim to mislead the attacker and divert them from their intended direction. To achieve this, various techniques are employed to modify the input data or manipulate the model's internal workings. For instance, some defenses may introduce random noise or distortion to the input samples, making them less susceptible to adversarial perturbations. This noise acts as a smokescreen, confusing the attacker and making it harder for them to generate effective adversarial examples.

We list the defensive methods evaluated in this paper below:

\textbf{Random Noise Defense (RND)~\cite{qin2021random}} is a lightweight defense that adds Gaussian noise to the input for each query. This work also theoretically shows RND's effectiveness against query-based attacks. 

\textbf{Small Noise Defense (SND)~\cite{Byun2021OnTE}} is also a randomized defense that uses a small additive input noise to neutralize
query-based attacks.

\textbf{Adversarial Attack on Attackers (AAA)~\cite{chen2022aaa}} directly optimizes the model's logits to confound the attacker towards incorrect attack directions.

\section{Method}\label{main_part}
\subsection{Problem Formulations}
\textbf{Adversarial attack.} Let $f:\mathbb{R}^d\to \mathbb{R}^K$ be the victim model, where $d$ is the input dimension, $K$ is the number of classes, $f_k(x)$ is the predicted score of class $k$ for input $x$. Given an input example $(x, y)$, the goal of adversarial attack is to find a sample $x'$ such that
\begin{equation}
    \arg\max_kf(x') \neq y,\quad \text{s.t}\quad d(x, x') \leq\epsilon,
\end{equation}
where $d(x, x')$ is distance between samples $x$ and $x'$. In practice, the distance can be the $\ell^2-$norm, $\|x-x'\|_2$, or the $\ell^{\infty}-$norm, $\|x-x'\|_{\infty}$.

This adversarial task can be framed as a constrained optimization problem. More particularly, the attacker tries to solve the following objective 
\begin{equation}
    \min_{x'}\mathcal{L}(f(x'), y),\quad \text{s.t}\quad d(x, x') \leq\epsilon,
\end{equation}
where $\mathcal{L}(.,.)$ is a loss function designed by the attacker. In practice, a common loss function $\mathcal{L}$ is the max-margin loss, as follows:
\begin{equation}
    \mathcal{L}(f(x), y)=f_y(x) - \max_{i\neq y}f_i(x).
\end{equation}
\textbf{Score-based attack.} For the query-based attack, an attacker can only access the input and output of the model; thus, the attacker cannot compute the gradient of the objective function with respect to the input $x$. However, the attacker can approximate the gradient using the finite difference method:
\begin{equation}
    \hat\nabla \mathcal{L}=\sum_{u}\frac{\mathcal{L}(f(x+\eta u), y)-\mathcal{L}(f(x), y)}{\eta}u, \quad\text{where } u\sim\mathcal{N}(0, \mu I).
\end{equation}

% {\color{red} more on score based atk}

Another approach to minimize the objective function is via random search. Specifically, the attacker proposes an update $u$ and computes the value of $\Lcal$ of this update to determine if $u$ can help improve the value of the objective function. Formally, the proposed $u$ is selected if $\Lcal(f(x+u), y) - \Lcal(f(x), y)<0$, otherwise it is rejected.

\textbf{Decision-based attack.} In contrast to score-based attacks, hard-label attacks find the direction that has the shortest distance to the decision boundary. The objective function of an untargeted hard-label attack can be formulated as follows:
\[\min_d g(d)\quad\]
\begin{equation}
    \text{where}\quad g(d)=\min \big\{r: \arg\max_kf(x+rd/\|d\|_2)\neq y\big\}.
\end{equation}
This objective function can be minimized using binary search, in which the attacker queries the model to find the distance $r$ for a particular direction $d$. To improve the querying efficiency, binary search can be combined with fine-grained search, in which the radius is iteratively increased  until the attacker finds an interval that contains $g(d)$. Hence, the gradient of $g(d)$ can also be approximated by the finite difference method
\begin{equation}
    \hat\nabla g(d)=\sum_u \frac{g(d+\eta u) - g(d)}{\eta}u.
\end{equation}
% {\color{red} more on decision based atk}

Similar to the case of score-based attacks, the attacker can also search for the optimal direction. Given the current best distance $r_{\operatorname{opt}}$, a proposed direction $d$ is eliminated if it cannot flip the prediction using the current best distance $r_{\operatorname{opt}}$; otherwise the binary search is used to compute $g(d)$, which is the new best distance.

\textbf{Randomized model.} In this work, we consider a randomized model $\randf: \Rbb^d\to \mathcal{P}(\Rbb^K)$ that maps a sample $x\in\Rbb^d$ to a probability distribution on $\Rbb^K$. Given an input $x$ and an attack query, the corresponding output is a vector drawn from $\randf(x)$. We assume that the randomized model $\randf$ is 'nice'; that is, the mean and variance of $\randf(x)$ exist for every $x$.

Finally, we define adversarial samples for a randomized model. Since the model has stochasticity, the prediction returned by the model of a sample $x$ can be inconsistent at different queries; i.e., the same sample can be correctly predicted at one application of $\randf$ and be incorrectly predicted later in another application of $\randf$. For this reason, adversarial attacks are successful if the obtained adversarial example can fool the randomized model in the majority of its applications on the example. 

\begin{definition}[Attack Success on Randomized Model]\label{adv_label}
    Given a datapoint $x$ with label $y$ and a positive real number $\epsilon$, a point $x'$ is called adversarial samples in a closed ball of radius $\epsilon$ around $x$ with respect to the model $\randf$ if $\|x'-x\|_p<\epsilon$ and
    \[\arg\max\Ebb [\randf(x')]\neq y.\]
\end{definition}
% \tung{The above formula does not include $x^{\prime}$}
\subsection{Randomized Feature Defense}\label{subsec:method}
\begin{wrapfigure}{R}{0.45\textwidth}
\vspace{-23pt}
\begin{minipage}{0.45\textwidth}
\begin{algorithm}[H]
\caption{Randomized Feature Defense}\label{alg:main}
\begin{algorithmic}
\Require a model $f$, input data $x$, 
\\noise statistics $\Sigma$, a set of perturbed layers \\$H=\{h_{l_0},h_{l_1},\dots,h_{l_n}\}$
\Ensure logit vector $l$ %\hl{is it f?}
\State $z_0\gets x$
\For {layer $h_i$  in the model}
\If{$h_i\in H$}
\State $\delta \sim \mathcal{N}(0, \Sigma)$
\State $z_i \gets h_i(z_{i-1}) + \delta$
\EndIf
\EndFor
\State $l\gets z_n$
\end{algorithmic}
\end{algorithm}
\end{minipage}
\vspace{-5pt}
\end{wrapfigure}
Our method is based on the assumption that the attacker relies on the model's output to find the update vector toward an adversarial example. Consequently, if the attacker receives unreliable feedback from the model, it will be more challenging for the attacker to infer good search directions toward the adversarial sample.

In contrast to the previous inference-time randomization approaches,  we introduce stochasticity to the model by perturbing the hidden features of the model. Formally, let $h_{l}$ be the $l-$th layer of the model, we sample an independent noise vector $\delta$ and forward $h_l(x)+\delta$ to the next layer. For simplicity, $\delta$ is sampled from Gaussian distribution $\Ncal(0, \Sigma)$, where $\Sigma$ is a diagonal matrix, or $\Ncal(0, \nu I), \nu \in \Rbb$. The detailed algorithm is presented in Algorithm~\ref{alg:main}.

Let $\randf$ be the proposed randomized model corresponding to the original $f$. When the variance of injected noise is small, we can assume that small noise diffuses but does not shift the prediction.
\begin{assumption}
    Mean of the randomized model $f_{\operatorname{rand}}$ with input $x$ is exactly the prediction of the original model for $x$
    \[\Ebb[f_{\operatorname{rand}}(x)]=f(x).\]
\end{assumption}
By Definition~\ref{adv_label}, adversarial samples of the original model are adversarial samples of the randomized model. Therefore, the direction that the attacker seeks is also that of the original model. Recall that the attacker finds this direction by either finite difference or random search.

In our method, when the model is injected with an independent noise, the value of objective $\Lcal$ is affected. If $\Lcal(f_{\mathrm{rand}}(x+\eta u), y)-\Lcal(f_{\mathrm{rand}}(x), y)$ oscillates among applications of $\randf$, the attacker is likely misled and selects a wrong direction. For random-search attacks, when the sign of $\Lcal(f_{\mathrm{rand}}(x+\eta u), y)-\Lcal(f_{\mathrm{rand}}(x), y)$ and the sign of $\Lcal(f(x+\eta u), y)-\Lcal(f(x), y)$ are different, the attacker chooses the opposite action to the optimal one. In other words, the attacker can either accept a bad update or reject a good one in a random search.

\subsection{Robustness to Score-based Attacks}\label{sec:score}

In this section, we present the theoretical analysis of the proposed defense against score-based attacks.

% \begin{theorem}\label{thm:robustness}
\begin{restatable}{thm}{robustness}\label{thm:robustness}
    Assuming the proposed random vector $u$ is sampled from a Gaussian $\Ncal(0, \mu I)$, the model is decomposed into $f=g\circ h$, and the defense adds a random noise $\delta\sim\Ncal(0, \nu I)$ to the output of $h$. At input $x$, the probability that the attacker chooses an opposite action positively correlates with \[\arctan\left(-\left(\frac{2\nu}{\mu}\frac{\|\nabla_{h(x)} (\mathcal{L}\circ g)\|^2_2}{\|\nabla_x (\mathcal{L}\circ f)\|^2_2}\right)^{-0.5}\right).\]
\end{restatable}
% \end{theorem}

This theorem states that the robustness of the randomized model is controlled by both (i) the ratio between the defense and attack noises and (ii) the ratio of the norm of the gradient with respect to the feature $h(x)$ and the norm of the gradient with respect to the input $x$. Since $\arctan$ is monotonically increasing, the model becomes more robust if the ratio $\frac{2\nu}{\mu}\frac{\|\nabla_{h(x)} (\mathcal{L}\circ g)\|^2_2}{\|\nabla_x (\mathcal{L}\circ f)\|^2_2}$ is high. Intuitively, the perturbations added by the attacker and by the defense induce a corresponding noise in the output; if the attack noise is dominated by the defense noise, the attacker cannot perceive how its update affects the model. Note that the $\arctan$ function is bounded, which means at some point the robustness saturates when the ratio increases.

\begin{figure}
    \centering
    \includegraphics[width=\linewidth]{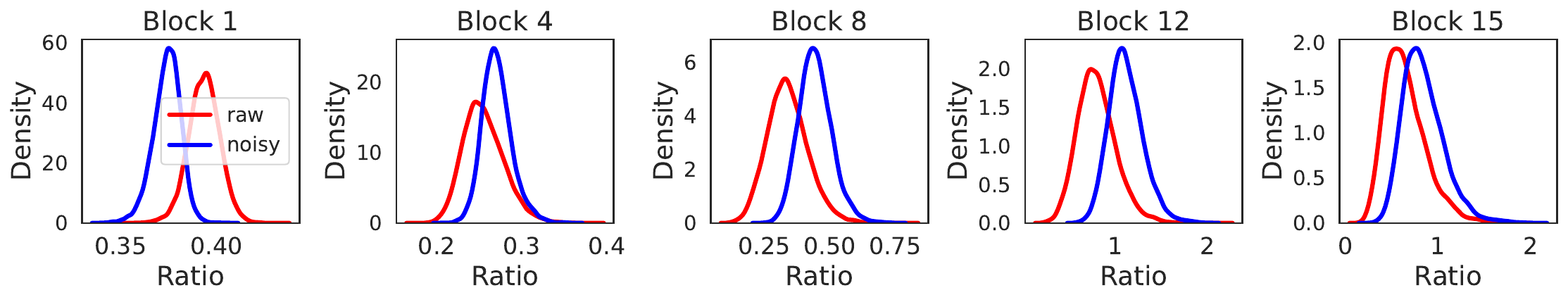}
    % \caption{The ratio of the norm of the gradient of $\Lcal$ at different layers and at input of VGG19 on CIFAR10 before and after perturbed.}
    \vspace{-20pt}
    \caption{The ratio of the norm of the gradient of $\Lcal$ at selected hidden layers and at input of VGG19 on CIFAR10 before and after perturbed. Full results are provided in the supplementary material.}
    \label{fig:vgg_ratio}
    \vspace{-5pt}
\end{figure}

While the first ratio is predetermined before an attack, the second ratio varies during the attack when the input $x$ is sequentially perturbed since it depends on the gradient of the objective function. To understand this behavior of the randomized model during the attack, we perform the following experiment. First, we compute the ratio of the norms of gradients at $h(x)$ and $x$. To simulate an attacker, we perform a single gradient descent step with respect to $\Lcal$. The distributions of the ratios on the raw and perturbed images at different layers are shown in Figure~\ref{fig:vgg_ratio}. We can observe that these ratios become higher when the data are perturbed toward the adversarial samples. In other words, 
the randomized model is more robust during the attack.

\vspace{-5pt}
\subsection{Robustness to Decision-based Attacks}\label{sec:decision}
\vspace{-5pt}
In decision-based attacks, the attacker finds the optimal direction $d_{\operatorname{opt}}$ and the corresponding distance $\ropt$ to the decision boundary such that $\ropt$ is minimal. We use the objective function $\Lcal(f(x), y)$ to understand how our method affects the decision-based attacks. Indeed, $\Lcal$ measures how close the prediction is to the true label: $\Lcal\leq 0$ if the prediction is incorrect and $\Lcal>0$ otherwise. 

To estimate $g(d)$, the attacker can use binary search. Similar to score-based attacks, when noise is injected into the model, the function $g(d)$ becomes stochastic, which eventually affects the attack. Unfortunately, the distribution of $g(d)$ (under binary search with randomness) does not have an analytical form. Nevertheless, we can still use a similar analysis to the last section to understand the robustness of our method. 
\begin{figure}
    \centering
    \includegraphics[width=\linewidth]{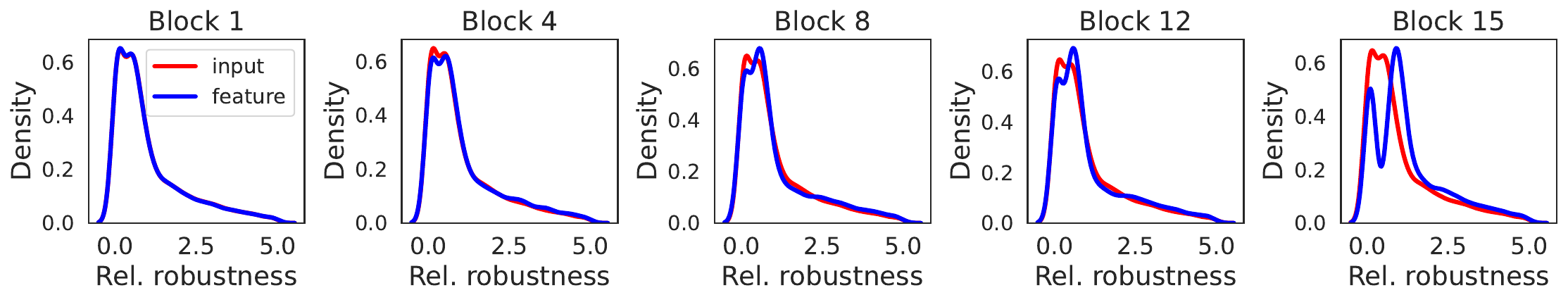}
    % \caption{The distribution of the magnitude of the robustness to query-based attacks of VGG19 computed at input and hidden feature on CIFAR10.}
    \vspace{-20pt}
    \caption{Distributions of the magnitude of the robustness to query-based attacks computed at input and selected hidden layers of VGG19 on CIFAR10. 
    % Full results are provided in the supplementary material.
    }
    
    \label{fig:vgg_acc}
    \vspace{-15pt}
\end{figure}
To avoid performing a binary search on uninformative directions, the attacker relies on best-radius searching. Given the current best distance $\ropt$, for every new direction $d$, the attacker verifies if the distance along $d$ to the boundary is shorter than $\ropt$ by querying $x+\ropt d/\|d\|_2$. When adding noise to features $h(x)$ of $f=g\circ h$ and linearizing the function at the current input $x$, we have 
\begin{align}
    \Lcal(\randf(x+\ropt d/||d||_2), y)&\approx \Lcal(g(h(x)+\ropt J_{h}(x)d/||d||_2 + \delta) \\
    &\approx \Lcal(f(x), y)  +\ropt\nabla_x\Lcal(f(x), y)d/||d||_2 + \nabla_{h(x)}\Lcal(g(h(x)), y)\delta \\
    &\approx (\ropt-g(d))\nabla_x\Lcal(f(x), y)d/||d||_2 + \nabla_{h(x)}\Lcal(g(h(x)), y)\delta,
\end{align}
% \tung{the first equation is not obvious to me}
where $J_h(x)$ is the Jacobian matrix of $h$ evaluated at $x$, since $\Lcal(f(x), y) + g(d)\nabla_x\Lcal(f(x), y)d/||d||_2\approx \Lcal(f(x+g(d)d/||d||_2), y)=0.$ If $\delta\sim\Ncal(0, \nu I)$, the variance of $\nabla_{h(x)}\Lcal(g(h(x)), y)\delta$ is $\nu\|\nabla_{h(x)}\Lcal(g(h(x)), y)\|_2^2$. When this value is large, it can dominate the other terms and increase the chance of flipping the sign of the loss function $\Lcal$. In other words, when $\Lcal$ has a high variance, the attacker is more likely to misjudge the direction.

\subsection{The Effect of Randomized Features on Accuracy}\label{sec:acc}
Let $\mathcal{D}$ be the data distribution, without any attack or defense, the accuracy of the model is
\begin{equation}
    \acc(f) := \Ebb_{(x,y)\sim\mathcal{D}}[\ind(f(x)=y)]=\Ebb_{(x,y)\sim\mathcal{D}}[\ind(\mathcal{L}(f(x),y)>0)].
\end{equation}
When injecting noise into the model, it becomes a robust, stochastic model $\randf:\Rbb^d\to\mathcal{P}(\Rbb^K)$. 

The clean accuracy of the randomized model is
\begin{equation}
    \acc(\randf)=\Ebb_{(x, y)\sim\mathcal{D}} \;\; \Ebb_{y'\sim \randf(x)}[\ind(y'=y)] \;=\Ebb_{(x, y)\sim\mathcal{D}} \;\; \Ebb_{y'\sim \randf(x)}[\ind(\mathcal{L}(y',y)>0)].
\end{equation}
% Consider the case where the defense injects noise $\delta_1\sim\mathcal{N}(0, \nu_1 I)$ to the input directly:
% \begin{align}
%     \acc(\randf)&=\Ebb_{(x, y)\sim\mathcal{D}} \;\; \Ebb_{\delta_1\sim\mathcal{N}(0, \nu_1 I)} \; [\ind(\mathcal{L}(f(x+\delta_1),y)>0)]\\
%     &\approx \Ebb_{(x, y)\sim\mathcal{D}} \;\; \Ebb_{\delta_1\sim\mathcal{N}(0, \nu_1 I)} \; [\ind(\mathcal{L}(f(x),y)+\nabla_x(\mathcal{L}\circ f)\delta_1>0)] \\
%     &=\Ebb_{(x, y)\sim\mathcal{D}} \;\; \Ebb_{\delta'_1\sim\mathcal{N}(0, \nu_1)} \; [\ind(\mathcal{L}(f(x),y)/\|\nabla_x(\mathcal{L}\circ f)\|_2+\delta'_1>0)].
% \end{align}
% \tung{In the last equation, the $\nabla_x$ disappears, why? We also need a product between $\nabla_x$ and $\delta_1$}
Adding noise $\delta_2\sim\mathcal{N}(0, \nu_2 I)$ to the features at layer $h$ of the model $f=g\circ h$ results in:
\begin{align}
    \acc(\randf)&=\Ebb\limits_{(x, y)\sim\mathcal{D}} \;\; \Ebb_{\delta\sim\mathcal{N}(0, \nu_2 I)} \; [\ind(\mathcal{L}(g(h(x)+\delta_2),y)>0)]\\
    &\approx \Ebb_{(x, y)\sim\mathcal{D}} \;\; \Ebb_{\delta_2\sim\mathcal{N}(0, \nu_2 I)} \; [\ind(\mathcal{L}(f(x),y)+\nabla_{h(x)}(\mathcal{L}\circ g)\delta_2>0)] \\
    &=\Ebb_{(x, y)\sim\mathcal{D}} \;\; \Ebb_{\delta'_2\sim\mathcal{N}(0, \nu_2)} \; [\ind(\mathcal{L}(f(x),y)/\|\nabla_{h(x)}(\mathcal{L}\circ g)\|_2+\delta'_2>0)].
\end{align}
It means that the accuracy of a randomized model depends on the objective function and its gradient, which vary for different data points. 
These ratios of $\Lcal$ and its gradient computed at the input and hidden layers are different. If $\Lcal$ is small at samples that have a large gradient norm when noise is injected at a layer, these samples will be likely misclassified while the correctly classified samples have a low magnitude of robustness (i.e., $\nu 
 \|\nabla_{h(x)} (\mathcal{L}\circ g)\|_2^2$ is small, as discussed in Theorem~\ref{thm:robustness} and Section~\ref{sec:decision}). In contrast, if the gradient norm with respect to the randomized layer is large for samples that have large $\Lcal$, the robustness of the model for the correctly classified samples will be high; thus, adding noise to this layer makes the model more robust against black-box attacks.
 
We conduct the following experiment to understand how the defense affects the whole dataset. We first compute the ratios of $\Lcal$ and its gradient for all samples and keep the top 99\% values. 
Essentially, the standard deviation of defensive noises that makes the accuracy drop by $1\%$ is proportional to the value at which $1\%$ of the ratios in the dataset are smaller. The product of this value and the norm of gradient represents the robustness of datasets, which are shown in Figure~\ref{fig:vgg_acc}.

We can observe that the ratio distributions  when randomizing the input and the hidden features are similar at the first few layers of the model; however, these ratios at the deeper layers of the model are higher. This means that randomizing the model at these layers makes it more robust than adding noise to the input layer when the defenders desire similar clean accuracy in the randomized models. 
\vspace{-15pt}
\section{Experiments}
\vspace{-5pt}
In this section, we evaluate the empirical performance of the proposed randomized feature defense.

\vspace{-5pt}
\subsection{Experimental Setup}
\vspace{-5pt}
\noindent \textbf{Datasets}. We perform our experiments on two widely used benchmark datasets in adversarial robustness: CIFAR10~\cite{cifar10} and ImageNet~\cite{ILSVRC15}. We randomly select $1000$ images that contain every class from the studied dataset in each experiment.

% \vspace{2pt}
\noindent \textbf{Defenses}. In addition to the proposed defense, we also include the related input defenses~\cite{qin2021random, Byun2021OnTE} in our evaluation. Note that, the empirical robustness comparison of all adversarial defenses is beyond the scope of the paper since our objective is to theoretically and empirically study the effectiveness of the randomized feature defense. We also evaluate AAA defense~\cite{chen2022aaa} against decision-based attacks and compare them with randomized defenses.

% \vspace{2pt}
\noindent \textbf{Attacks}. For score-based attacks, we consider the gradient-estimation methods, NES~\cite{ilyas2018black}, and the random-search methods, Square ~\cite{andriushchenko2020square}, SignHunt~\cite{al2020sign}. For decision-based attacks, we consider RayS~\cite{chen2020rays} and SignFlip~\cite{Chen2020boosting}.

% \vspace{2pt}
\noindent \textbf{Models}. We consider $6$ victim models on ImageNet, including $2$ convolution models that are VGG19~\cite{Simonyan2015vgg} and ResNet50~\cite{He2016resnet}, $2$ transformer models that are ViT~\cite{dosovitskiy2020image} and DeiT~\cite{touvron21a2021deit}. For the experiments on CIFAR10, we finetuned VGG19, ResNet50, ViT, DeiT with an input size of $224\times 224$.

% \vspace{5pt}
\noindent \textbf{Evaluation protocol}. For a fair comparison, we report each defense's robustness performance results at the  corresponding configuration of hyperparameters that achieves a specific drop (i.e., $\approx$1\% or $\approx$2\%) in clean-data accuracy. In practice, a defender always considers the trade-off between robustness and clean-data performance, with a priority on satisfactory clean-data performance; thus, achieving higher robustness but a significant drop in clean-data accuracy is usually not acceptable. 
\vspace{-5pt}
\subsection{Performance against Score-based Attacks}
\vspace{-5pt}
On ImageNet, we report the accuracy under the attack of $6$ models and $3$ score-based attacks in Table~\ref{tab:imagenet}. As we can observe, while the attacks achieve close to $0\%$ failure rate on the base models (i.e., without any defense), both randomized feature and input defenses significantly improve the models' robustness against score-based attacks. Furthermore, for Square attack and SignHunt, which are strong adversarial attack baselines, randomized feature defense consistently achieves better performance on all $6$ models, which supports our theoretical analysis in Section~\ref{main_part}. For instance, while the base VGG19 models are severely vulnerable, our randomized feature defense achieves $22.2\%$ in robust accuracy after $10000$ query, also significantly better than the randomized input defense ($17.8\%$ robust accuracy). On the transformer-based DeiT, our randomized feature defense has $69.1\%$ robust accuracy under Square attack, while the robust accuracy of the randomized input defense is $2\%$ lower. For the NES attack, the randomized-feature VGG19 shows the best robustness. In summary, randomized feature defense consistently achieves high robustness on most models except ResNet50 where the robustness is similar to randomized input defense. 
\vspace{-10pt}
\begin{table}[H]
\parbox[t]{.45\linewidth}{    
    \scriptsize
    \setlength{\tabcolsep}{1pt}
    \renewcommand{\arraystretch}{0.5}
    \centering
    \caption{Defense Performance in ImageNet. The clean-data accuracy of the robust models is allowed to drop either $\approx 1\%$ or $\approx 2\%$.}
    \vspace{-5pt}
    \mbox{\hspace{-15pt}
    \begin{tabular}{lllcccccc}
        \toprule
         \multirow{2}{*}{Model} & \multirow{2}{*}{Method}  & \multirow{2}{*}{Acc} & 
         \multicolumn{2}{c}{Square} &
        \multicolumn{2}{c}{NES} &
        \multicolumn{2}{c}{SignHunt} \\
        \cmidrule(l){4-5} \cmidrule(l){6-7} \cmidrule(l){8-9} 
        &&& 1000 & 10000 &1000 & 10000 & 1000 & 10000\\
        \midrule
        \multirow{5}{*}{ResNet50} & Base  & 80.37 & 3.5 & 0.2 & 36.2 & 4.3 & 6.6 & 0.4 \\
        \cmidrule{2-9}
        & \multirow{2}{*}{Input}  & 79.18 ($\approx 1\%$) & 40.3 & 39.5 & 63.8 & 23.9 & 47.6 & 45.4 \\
        &  & 78.46 ($\approx 2\%$) & 41.1 & 39.8 & \bf 69.4 & \bf 41.5 & 49.3 & 47.2 \\
        \cmidrule{2-9}
        & \multirow{2}{*}{Feature}  & 79.70 ($\approx 1\%$) & 37.0 & 36.0 & 56.7 & 16.8 & 46.3 & 43.4 \\
        &  & 78.43 ($\approx 2\%$) & \bf 42.0 & \bf 41.5 & 65.6 & 40.6 & \bf 51.3 & \bf 49.3\\
        \midrule
        \multirow{5}{*}{VGG19} & Base  & 74.21 & 0.1 & 0.0 & 19.6 & 0.0 & 0.4 & 0.0\\ 
        \cmidrule{2-9}
        & \multirow{2}{*}{Input}  & 73.24 ($\approx 1\%$) & 7.7 & 6.9 & 32.1 & 1.5 & 18.3 & 17.0 \\
        &  & 71.43 ($\approx 2\%$) & 18.7 & 17.8 & 47.4 & 11.5 & 28.3 & 27.1 \\
        \cmidrule{2-9}
        & \multirow{2}{*}{Feature}  & 72.66 ($\approx 1\%$) & 22.4 & 21.6 & 50.1 & 18.5 & 34.6 & \bf 32.9\\
        &  & 71.21 ($\approx 2\%$) & \bf 23.3 & \bf 22.2 & \bf 55.1 & \bf 28.4 & \bf 36.5 & 32.8 \\
        \midrule
        \multirow{5}{*}{DeiT} & Base  & 82.00 & 6.4 & 0.0 & 46.7 & 0.8 & 22.3 & 0.0 \\
        \cmidrule{2-9}
        & \multirow{2}{*}{Input}  & 80.10 ($\approx 1\%$) & 67.7 & 67.2 & \bf 75.8 & 65.9 & 64.4 & 63.6 \\
        &  & 79.60 ($\approx 2\%$) & 66.6 & 66.0 & 75.7 & \bf 67.1 & 64.9 & 64.3 \\
        \cmidrule{2-9}
        & \multirow{2}{*}{Feature}  & 80.80 ($\approx 1\%$) & \bf 69.7 & \bf 69.1 & 75.0 & 59.1 & \bf 66.4 & 64.1\\
        & & 79.76 ($\approx 2\%$) & 69.3 & 69.0 & 75.1 & 65.3 & 66 & \bf 64.3 \\
        \midrule
        \multirow{5}{*}{ViT} & Base  & 79.15 & 5.7 & 0.0& 45.7 & 7.3 & 5.1 & 0.0\\
        \cmidrule{2-9}
        & \multirow{2}{*}{Input}  & 78.28 ($\approx 1\%$) & 58.8 & 58.1 & 70.8 & 51.4 & 53.1 & 52.2 \\
        &  & 77.09 ($\approx 2\%$) & 61.3 & 60.9& 70.6 & \bf 59.2 & 53.7 & 52.7 \\
        \cmidrule{2-9}
        & \multirow{2}{*}{Feature}  &78.20 ($\approx 1\%$) & 60.6 & 60.2 & 69.1 & 47.5 & 54.0 & 52.9 \\
        &  & 77.18 ($\approx 2\%$) & \bf 63.7 & \bf 62.9   & \bf 72.2 & 58.1 & \bf 57.0 & \bf 55.3 \\

        \bottomrule
    \end{tabular}
    }
    \label{tab:imagenet}
}
\hfill
\parbox[t]{.45\linewidth}{
    % \vspace{-5pt}
    \scriptsize
    \setlength{\tabcolsep}{1pt}
    \renewcommand{\arraystretch}{0.5}
    \centering
    \caption{Defense Performance in CIFAR10. The clean-data accuracy of the robust models is allowed to drop either $\approx 2\%$ or $\approx 4\%$.}
    \vspace{-5pt}
    \mbox{\hspace{-15pt}
    \begin{tabular}{lllcccccc}
        \toprule
         \multirow{2}{*}{Model} & \multirow{2}{*}{Method}  & \multirow{2}{*}{Acc} & 
         \multicolumn{2}{c}{Square} &
        \multicolumn{2}{c}{NES} &
        \multicolumn{2}{c}{SignHunt} \\
        \cmidrule(l){4-5} \cmidrule(l){6-7} \cmidrule(l){8-9} 
        &&& 1000 & 10000 &1000 & 10000 & 1000 & 10000\\
        \midrule
         \multirow{5}{*}{ResNet50} & Base & 97.66 & 0.8 & 0.1 & 71.7 & 21.7 &  3.7 & 0.2\\
         \cmidrule{2-9}
         & \multirow{2}{*}{Input} & 95.98 ($\approx 2\%$) & 50.5 & 48.8 & 93.1 & 85.4 & 26.8 & 26 \\
 
         &  & 93.42 ($\approx 4\%$) & 56.4 & \bf 54.8 & 90.0 & 85.0 & 31.1 & 29.8 \\
         \cmidrule{2-9}
         & \multirow{2}{*}{Feature}  & 95.95 ($\approx 2\%$) & 54.9 & 52.8 & \bf 93.2 & \bf 86.2 & 32.5 & 30.6 \\

         &  & 93.48 ($\approx 4\%$) & \bf 56.7 & 53.4 & 89.9 & 83.9 & \bf 37.1 &  \bf 35.7\\
         \midrule
         \multirow{5}{*}{VGG19} & Base  & 96.28 & 0.6 & 0.1 & 68.8 & 16.6 & 3.2 & 0.3\\
         \cmidrule{2-9}
         & \multirow{2}{*}{Input} 

          & 94.92 ($\approx 2\%$) & 30.6 & 27.1 & 89.5 & 58.0 & 22.7 & 21.8 \\
         &  & 93.52 ($\approx 4\%$) & 42.2 & 39.8 & 90.3 & 68.4 & 27.5 & 26.8 \\
         \cmidrule{2-9}
         & \multirow{2}{*}{Feature}  &94.93 ($\approx 2\%$) & 61.0 & 58.4 & \bf 92.2 & 77.9 & 43.2 & 42.4 \\

         &  & 93.58 ($\approx 4\%$) & \bf 64.2 & \bf 62.8 & 91.2 & \bf 80.1 & \bf 49.2 & \bf 46.9 \\

         \midrule
         \multirow{5}{*}{DeiT} & Base & 98.40 & 3.2 & 0.0 & 81.9 & 34.2 & 7.9 & 0.2\\
         \cmidrule{2-9}
         & \multirow{2}{*}{Input}  & 96.59 ($\approx 2\%$) & 66.9 & 67.6 & \bf 95.2 & \bf 90.0 & 40.2 & 39.2 \\

         &  & 94.81 ($\approx 4\%$) & \bf 70.6 & \bf 68.8 & 92.6 & 87.7 & 40.3 & 38.5\\
         \cmidrule{2-9}

         & \multirow{2}{*}{Feature}  & 96.29 ($\approx 2\%$) & 69.1 & 67.9 & 94.1 & 88.3 & \bf 45.7 & \bf 43.4\\
         & & 94.91 ($\approx 4\%$) & 68.9 & 66.1 & 93.5 & 87.6 & 43.6 &40.4\\
         \midrule
         \multirow{5}{*}{ViT} & Base & 97.86 & 5.1 &0.0 & 84.8 & 43.6 & 6.1 & 0.0\\
         \cmidrule{2-9}
         & \multirow{2}{*}{Input} &  95.80 ($\approx 2\%$) & 63.0 & 61.2 & 93.5 & \bf 87.0 & 34.8 & 33.3 \\
         & &  93.40 ($\approx 4\%$) & 62.6 & 61.1 & 89.7 & 85.5 & 33.4 & 32.2\\
         \cmidrule{2-9}
         & \multirow{2}{*}{Feature} &  95.96 ($\approx 2\%$) & 63.9 & 62.7 & \bf 93.7 & 85.6 & 42.5 & 40.7 \\
         &  & 93.39 ($\approx 4\%$) &\bf  66.2 & \bf 65.6 & 92.9 & 85.3 & \bf 44.8 & \bf 43.8 \\
         \bottomrule
    \end{tabular}
    }
    \label{tab:cifar10}
}
% \vspace{-20pt}
\end{table}
\vspace{-10pt}
We also observe similar robustness results on CIFAR10 experiments with ResNet50, VGG19, DeiT, and ViT for $3$ attacks. As we can observe in Table~\ref{tab:cifar10}, randomized feature and input defenses are effective against score-based attacks. Similar to ImageNet, randomized feature defense achieves significantly better robustness than randomized input defense in most experiments. For Square attacks on ResNet50 and DeiT, while the best robustness is achieved by randomized input defense, randomized feature defense is more robust when the defender sacrifices $2\%$ clean-data accuracy.

% \begin{table}[]
\begin{wraptable}{l}{0.52 \textwidth}
\vspace{-10pt}
\setlength{\tabcolsep}{1pt}
\renewcommand{\arraystretch}{0.5}
\scriptsize
\centering
\caption{Robustness (higher means more robust) under different values of $\mu$. \textbf{Small $\nu$} corresponds to selected $\nu$ where clean accuracy is allowed to drop by $2\%$, and \textbf{Large $\nu$} corresponds to clean accuracy drop of $4\%$.}
% \vspace{-5pt}
\begin{tabular}{@{}llcccccccc@{}}
\toprule
\multirow{3}{*}{Attack} &
  \multirow{3}{*}{$\mu$} &
  \multicolumn{4}{c}{VGG} &
  \multicolumn{4}{c}{ViT} \\
  \cmidrule(l){3-6} \cmidrule(l){7-10}
 &
   &
  \multicolumn{2}{c}{Small $\nu$} &
  \multicolumn{2}{c}{Large $\nu$} &
  \multicolumn{2}{c}{Small $\nu$} &
  \multicolumn{2}{c}{Large $\nu$} \\
  \cmidrule(l){3-4} \cmidrule(l){5-6}  \cmidrule(l){7-8} \cmidrule(l){9-10}  
  
                          &       & Input & Feature & Input & Feature & Input & Feature & Input & Feature \\
                          \midrule
\multirow{4}{*}{Square}   & 0.05  & 30.6  & 61.0    & 42.2  & 64.2    & 63.0  & 63.9    & 62.6  & 66.2    \\
                          & 0.1   & 47.4  & 65.8    & 54.6  & 65.5    & 69.3  & 70.2    & 68.8  & 69.6    \\
                          & 0.2   & 32.1  & 59.7    & 43.9  & 64.0    & 56.1  & 58.0    & 56.8  & 58.6    \\
                          & 0.3   & 27.0  & 54.9    & 38.1  & 59.7    & 47.1  & 51.9    & 47.7  & 50.4    \\
                          \midrule
\multirow{4}{*}{NES}      & 0.001 & 93.4  & 93.9    & 90.1  & 91.4    & 93.7  & 94.8    & 90.3  & 93.5    \\
                          & 0.01  & 89.5  & 92.2    & 90.3  & 91.2    & 93.5  & 93.7    & 89.7  & 92.9    \\
                          & 0.1   & 88.0  & 90.0    & 86.7  & 89.6    & 87.9  & 91.4    & 86.7  & 90.6    \\
                          & 0.2   & 93.6  & 93.0    & 92.6  & 91.4    & 91.0  & 93.8    & 87.6  & 92.0    \\
                          \midrule
\multirow{4}{*}{SignHunt} & 0.01  & 91.6  & 91.0    & 91.3  & 88.0    & 89.1  & 90.9    & 85.4  & 91.3    \\
                          & 0.05  & 22.7  & 43.2    & 27.5  & 49.2    & 34.8  & 42.5    & 33.4  & 44.8    \\
                          & 0.075 & 5.6   & 19.7    & 8.1   & 25.6    & 13.6  & 22.5    & 13.7  & 24.3    \\
                          & 0.1   & 1.2   & 7.9     & 2.4   & 12.1    & 5.5   & 11.3    & 5.2   & 12.7   \\ \bottomrule
\end{tabular}
\label{tab:varying_atk}
% \end{table}
\vspace{-15pt}
\end{wraptable}

\noindent \textbf{Dynamic Analysis of Robustness.}
As the adversary increases the magnitude of perturbation, the attack becomes more effective since the misleading probability decreases as shown in Theorem~\ref{thm:robustness}. The adversary can vary the square size for Square attack, the exploration step  for NES, and the budget for SignHunt (since SignHunt sets the finite-difference probe to the perturbation bound). 

Table~\ref{tab:varying_atk} reports the robustness of the models under stronger attacks from these adversaries for  different values of $\nu$. 
We can observe that increasing the strength of the attack leads to lower robustness among all the defenses. However, at the selected defense noise scales corresponding to the same clean accuracy drop, our defense is still more robust than randomized input defense; this improved robustness again can be explained by the analysis in Section~\ref{sec:score} and \ref{sec:acc}. A larger attack perturbation may also cause the approximation in the attack to be less accurate, which leads to a drop in the attack’s effectiveness; for example, the robustness increases from $89.6\%$ to $91.4\%$ when the NES’s perturbation magnitude increases in VGG19 experiments (similar observations in ViT).

\noindent \textbf{Combined with Adversarial Training (AT).} We evaluate the combination of our defense and AT on CIFAR10/ResNet20 model against under score-based attacks with 1000 queries and observe significantly improved robustness, as shown in Table~\ref{tab:at}.
\vspace{-5pt}
\subsection{Performance against Decision-based Attacks}
\vspace{-5pt}

Table~\ref{tab:cifar10_decision} reports the performance of VGG19 and ResNet50 against $2$ decision-based attacks on CIFAR10. Besides randomized feature and input defenses, we also include AAA defense, which optimizes the perturbation that does not change the prediction. While AAA is optimized for score-based attacks directly and thus is successful in fooling these attacks (as seen in Table 3 in Supplementary), the results show that AAA is not effective in defending against decision-based attacks, while randomized feature and input defenses improve the robustness. An interesting observation is that RayS attack is more effective than score-based attacks although it only uses hard labels, even when there are defenses.
\vspace{-5pt}

\subsection{Relationship Between the Gradient Norm and the Robustness to Score-Based Attacks}
\vspace{-5pt}

In Table~\ref{tab:layerwise_fix_scale}, we provide the corresponding accuracy under attack on CIFAR10 with 1000 queries (for when a single layer is randomized with a fixed value of $\nu$) and the mean of the gradient norm at that layer. As we can observe, as the gradient norm increases (also as we originally observe in Figure~\ref{fig:vgg_ratio}), the robustness also increases, thus verifying our theoretical results.

\begin{table}[t!]
\parbox[]{.25\textwidth} {
\setlength{\tabcolsep}{1pt}
    \scriptsize
    \centering
    \caption{Robustness with adversarial training.}
    \vspace{-5pt}
    \begin{tabular}{@{}lccc@{}}
    \toprule
            & Square & NES  & SignHunt \\ \midrule
    AT      & 32.5   & 67.6 & 31.7     \\
    Ours    & 37.6   & 44.1 & 41.7     \\
    Ours+AT & 77.8   & 80.6 & 67.0  \\
    \bottomrule
    \end{tabular}
    \label{tab:at}    
}
\hfill
\parbox[]{.35\textwidth} {
 \setlength{\tabcolsep}{1pt}
 \renewcommand{\arraystretch}{0.7}
    \scriptsize
    \centering
    \caption{Robustness against decision-based attacks (CIFAR10)}
    % \vspace{-2pt}
    \vspace{-5pt}
    \begin{tabular}{lllcc}
        \toprule
         Model & Method & Acc & RayS & SignFlip \\
         \midrule 
        \multirow{4}{*}{ResNet50} & Base & 97.66 & 0.1 & 20.5  \\
        & AAA & 97.70 & 0.1 & 20.4  \\
        & Input & 93.52 & 12.0 & \bf 85.5  \\
        & Feature & 92.10 & \bf 14.4 & 82.5 \\
        \midrule
        \multirow{4}{*}{VGG19} & Base & 96.28 & 0.0 & 6.4  \\
        & AAA & 96.30 & 0.1 & 5.7   \\
        & Input & 93.42 & 8.1 & \bf 86.0  \\
        & Feature & 93.48 & \bf 15.4 & 76.5  \\
        
        \bottomrule
         
    \end{tabular}
    
    \label{tab:cifar10_decision}
}
\hfill
\parbox[]{.35\textwidth} {
\setlength{\tabcolsep}{1pt}
\renewcommand{\arraystretch}{0.7}
\scriptsize
\centering
\caption{Robustness in CIFAR10 at each layer (fixed $\nu$).}
\vspace{-5pt}
\begin{tabular}{@{}llcccc@{}}
\toprule
Model & Layer & Square & NES  & SignHunt & GradNorm \\ \midrule
VGG   & 1     & 56.7   & 87.8 & 21.5     & 1.324   \\
      & 4     & 52.5   & 84.2 & 18.7     & 0.842   \\
      & 12    & 63.0   & 89.7 & 29.4     & 2.514   \\
      & 15    & 50.6   & 87.7 & 37.4     & 1.710   \\ \midrule
ViT   & 1     & 77.3   & 94.8 & 26.1     & 0.615   \\
      & 4     & 75.3   & 94.4 & 28.0     & 0.462   \\
      & 8     & 65.8   & 91.6 & 26.9     & 0.324   \\
      & 11    & 48.3   & 86.5 & 23.1     & 0.214   \\ \bottomrule
\end{tabular}
\label{tab:layerwise_fix_scale}
}
\vspace{-5pt}
\end{table}

\vspace{-5pt}
\subsection{Performance against Adaptive Attacks}
\vspace{-5pt}
We conduct experiments with adaptive attacks that apply Expectation Over Transformation (EOT)~\cite{athalye2017synthesizing} in which the attacker queries a sample $M$ times and averages the outputs to cancel the randomness. Tables~\ref{tab:adaptive} show the robust accuracy of VGG19 and ResNet50 on CIFAR10 against EOT attack with $M=5$ and $M=10$. Note that with EOT, the number of updates in the attack is $M$ times less than that of a normal attack with the same query budget. For this reason, we report the results for adaptive attacks with both $1000$ queries and $M\times 1000$ queries. We can observe that EOT can mitigate the effect of randomized defenses even with the same number of queries; however, feature defense still yields better performance.
\begin{table*}[t!]
\scriptsize
\setlength{\tabcolsep}{1pt}
\renewcommand{\arraystretch}{0.5}
\centering
\caption{Defenses against adaptive attacks on CIFAR10}
\vspace{-5pt}
\begin{tabular}{@{}llcccccccccccc@{}}
\toprule
\multirow{3}{*}{Attacks} & \multirow{3}{*}{Methods} & \multicolumn{6}{c}{VGG19} & \multicolumn{6}{c}{ResNet50} \\
\cmidrule(l){3-8} \cmidrule(l){9-14} 
 &  & \multirow{2}{*}{Acc} & $M=1$ & \multicolumn{2}{c}{$M=5$} & \multicolumn{2}{c}{$M=10$} & \multirow{2}{*}{Acc} & $M=1$ & \multicolumn{2}{c}{$M=5$} & \multicolumn{2}{c}{$M=10$} \\
  \cmidrule(l){4-4} \cmidrule(l){5-6} \cmidrule(l){7-8} \cmidrule(l){10-10} \cmidrule(l){11-12} \cmidrule(l){13-14}
 &  &  & QC=1000 & QC=1000 & QC=5000 & QC=1000 & QC=10000 &  & QC=1000 & QC=1000 & QC=5000 & QC=1000 & QC=10000 \\ \midrule
\multirow{2}{*}{Square } & Input & 94.92 &  30.6 & 24.2 & 10.5 & 30.2 & 3.2 & 95.32 &52.9 & 42.0 & 34.8 & 35.0 & 13.3  \\ 
 & Feature & 94.93 &  61.0 & 53.0 & 45.5 & 46.7 & 23.1 & 95.21 & 54.5 & 45.1 & 40.4 & 37.3 & 21.1  \\ \midrule
\multirow{2}{*}{NES} & Input & 94.92 &  89.5 & 93.4 & 82.1 & 94.4 & 78.8 & 95.32 & 92.4 & 94.0 & 91.3 & 93.9 & 90.7 \\
 & Feature & 94.93 & 92.2 & 94.8 & 88.4 & 94.5 & 86.0 & 95.21 &91.8 & 93.8 & 90.8 & 94.0 & 90.4 \\ \midrule
\multirow{2}{*}{SignHunt} & Input & 94.92 &  22.7 & 15.9 & 10.4 & 23.3 & 7.6 & 95.32 &29.9 & 17.6 & 13.5 & 21.1 & 9.4 \\
 & Feature & 94.93 &43.2  & 27.1 & 23.0 & 31.7 & 17.0 & 95.21 &35.1 & 17.3 & 16.4 & 21.5 & 11.3 \\ \bottomrule
\end{tabular}
\label{tab:adaptive}
\vspace{-15pt}
\end{table*}

\vspace{-5pt}
\section{Conclusion and Future Work}
\vspace{-5pt}
In this work, we study the effectiveness of random feature defense against query-based attacks, including score-based and decision-based attacks. We provide an analysis that connects the robustness to the variance of noise and the local behavior of the model. Our empirical results show that random defense helps improve the performance of the model under query-based attacks with a trade-off in clean accuracy. Future works will be directed toward the analysis covering black-box attacks that transfer adversarial samples from the surrogate model to the target model.
%, {\color{red} or extending our defense to the white-box setting, which poses a challenge for our and the related randomized defenses since it is not straightforward how to intervene this attacking process}.

\bibliographystyle{iclr2024_conference}
\bibliography{iclr2024_conference, nips, iccv}
\newpage
% \appendix
% \section*{Appendix}
% \input{supp}

\end{document}